# NeuroEvo: A Cloud-based Platform for Automated Design and Training of Neural Networks using Evolutionary and Particle Swarm Algorithms

**Philip Schroeder**
Johns Hopkins University
pschroe9@jhu.edu

## Abstract

Evolutionary algorithms (EAs) provide unique advantages for optimizing neural networks in complex search spaces. This paper introduces a new web platform, NeuroEvo (neuroevo.io), that allows users to interactively design and train neural network classifiers using evolutionary and particle swarm algorithms. The classification problem and training data are provided by the user and, upon completion of the training process, the best classifier is made available to download and implement in Python, Java, and JavaScript. NeuroEvo is a cloud-based application that leverages GPU parallelization to improve the speed with which the independent evolutionary steps, such as mutation, crossover, and fitness evaluation, are executed across the population. This paper outlines the training algorithms and opportunities for users to specify design decisions and hyperparameter settings. The algorithms described in this paper are also made available as a Python package, neuroevo (PyPI: https://pypi.org/project/neuroevo/).

## 1 Introduction

Evolutionary algorithms (EAs) are a class of problem solving and optimization methods inspired by mechanisms of biological evolution, including reproduction, mutation, recombination, and selection. Neural networks comprise one of many classes of models that can be effectively optimized with EAs for complex problems. EAs have several advantages over other methods of training neural networks. To start, unlike gradient-based methods, EAs do not require specific knowledge of the search space (e.g., first or second derivatives, discontinuities, etc.). EAs simply require that the relative fitness of each candidate solution be evaluated in accordance with the objective function. In addition, compared to deterministic methods, EAs are less susceptible to converging on local optima due to their capacity to simultaneously explore uncharted regions of the search space while exploiting profitable subregions that have already been discovered [1-5]. EAs are also well-suited for non-stationary search spaces, where solutions must be dynamically re-adapted in response to variation that occurs in the problem environment [6-7]. Finally, due to the decentralized nature of EAs, their time efficiency can be significantly improved through parallelization, with serial processing only required during the selection process [8-11].

This paper introduces a web platform, NeuroEvo ([neuroevo.io](neuroevo.io)), that allows users to interactively design and train neural network classifiers using evolutionary and particle swarm algorithms. Specifically, the platform allows users to design a multilayer perceptron (MLP) and train it using particle swarm optimization (PSO), differential evolution (DE), or a genetic algorithm (GA). The classification problem and training data are provided by the user. Once the training is complete, the classifier is made available to download and implement in Python, Java, and JavaScript. The following sections of this paper provide a description of the training algorithms and opportunities for users to specify design decisions and hyperparameter settings.

NeuroEvo is a cloud-based platform, built with Amazon Web Services (AWS) and funded by AWS research credits. The algorithms described in this paper are also made available with a lightweight NumPy-based Python package, neuroevo (PyPI: https://pypi.org/project/neuroevo/, GitHub: https://github.com/NeuroEvo/neuroevo).

## 2 Particle Swarm Optimization

The PSO algorithm is represented in Algorithm 1, where $\mathbf{U}(0, \phi)$ is a vector of random numbers uniformly distributed in $[0, \phi]$, which is generated for each particle for each iteration, and $\otimes$ is the component-wide multiplication. The algorithm begins by initializing a swarm of $NP$ candidate neural networks, $\Psi_{NP}^{(0)}$, with a random position, $\theta_i^{(0)}$, and velocity, $v_i$, within the search space. The structure of each MLP ("particle") is defined by the user-provided design parameters, including

---

**Algorithm 1**: Particle Swarm Optimization

**procedure PSO**

    Initialize population as $\Psi_{NP}^{(0)} = \{\theta_1^{(0)}, \theta_2^{(0)}, \dots, \theta_{NP}^{(0)}\}$ with random weights & velocity

    $t = 0$

    **while** $f(\theta_{best}) < K$ **and** $t < T_{max}$ **do**

        **for all** $\theta_i \in \Psi_{NP}^{(t)}$ **do**

$$\omega_i(t) = \omega(0) + (\omega(T_{max}) - \omega(0)) \frac{e^{m_i(t-1)} - 1}{e^{m_i(t-1)} + 1}$$

$$v_i = \omega(t) v_i + \mathbf{U}(0, \phi_i) \otimes (p_i - \theta_i^{(t)}) + \mathbf{U}(0, \phi_g) \otimes (p_g - \theta_i^{(t)})$$

$$\theta_i^{(t)} = \theta_i^{(t)} + v_i$$

            **if** $f\left(\theta_i^{(t)}\right) > f(p_g)$ **do**

$$p_g = \theta_i^{(t)}$$

            **end if**

            **if** $f\left(\theta_i^{(t)}\right) > f(p_i)$ **do**

$$p_i = \theta_i^{(t)}$$

            **end if**

$$m_i(t+1) = \frac{f(p_i) - f\left(\theta_i^{(t)}\right)}{f(p_i) + f\left(\theta_i^{(t)}\right)}$$

        **end for**

        $t \mathrel{+}= 1$

    **end while**

**end procedure**

the activation function, the number of hidden layers, and the number nodes for each hidden layer. Each particle represents a candidate solution to the problem and is coded as a vector. Upon every iteration of the algorithm, the current position of each particle is evaluated against the fitness function, $f\left(\theta_i^{(t)}\right)$. The best position for particle $i$ is stored in a vector $p_i$ (personal best) and the position of the particle with the best global fitness is stored in a vector $p_g$ (global best). Each iteration, the particle's velocity is updated according to the influence of the local best position, $p_i$, and the global best position, $p_g$. The degree of influence from $p_i$ and $p_g$ can be controlled with the user-provided parameters, $\phi_i$ and $\phi_g$, respectively. The new velocity is then used to update each particle's position. This algorithm continues until the user-determined satisfactory fitness level, $K$, is achieved or the maximum number of training iterations, $T_{max}$, is reached.

The inertial weight parameter, $w$, can be used to tune the balance of exploration and exploitation in the PSO search process and to control the swarm's rate of convergence. Users have the option of setting an inertial weight that is constant or dynamically adjusted using either a linear or nonlinear decreasing approach.

Linear decreasing: For this approach, $w$ starts with a large value, $\omega(0)$), and, on each iteration, $t$, decreases toward a smaller value, $\omega(T_{max})$, in a linear fashion according to the following function:

$$\omega(t) = \left(\omega(0) - \omega(T_{max})\right)\frac{T_{max} - t}{T_{max}} + \omega(T_{max}) \tag{1}$$

This requires $\omega(0) > \omega(T_{max})$. The default setting for $\omega(0)$ and $\omega(T_{max})$ is 0.9 and 0.5, respectively.

Nonlinear decreasing: This approach is similar to that above, but instead implements a nonlinear decreasing function:

$$\omega_i(t) = \omega(0) + \left(\omega(T_{max}) - \omega(0)\right)\frac{e^{m_i(t-1)} - 1}{e^{m_i(t-1)} + 1} \tag{2}$$

where the relative improvement $m_i(t-1)$ is defined as:

$$m_i(t-1) = \frac{f(p_i) - f\left(\theta_i^{(t-1)}\right)}{f(p_i) + f\left(\theta_i^{(t-1)}\right)} \tag{3}$$

with $\omega(T_{max}) \approx 0.5$ and $\omega(0) < 1$.

## 3  Differential Evolution

The DE algorithm is represented in Algorithm 2. Similar to the PSO algorithm, DE starts with a parent generation, a population of $NP$ candidate solutions, $\Psi_{NP}^{(0)}$, that are initialized using random weights. An evolutionary sequence is then mimicked by producing progeny candidates through a recombination process, inspired by the crossover events that occur during meiotic prophase I of germline cell division. Every candidate solution produces an offspring (trial vector), $U_i^{(t)}$, by crossing over with a donor vector, $V_i^{(t)}$, which is generated based on the operation described below. The degree of crossover is defined by the user-provided parameter $CR \in (0,1]$. Using the traditional binary crossover strategy, $CR$ represents the probability of the trial vector inheriting an element form the donor vector, as represented below in Equation (4).

$$U_i^{(t)} = \begin{cases} V_i^{(t)}, \text{if } rand(0,1) \leq CR \\ \theta_i^{(t)}, \text{otherwise} \end{cases} \quad (4)$$

If the trial vector is more fit than the corresponding parent candidate (target vector), $\theta_i^{(t)}$, then the trial vector replaces the target vector within the population. Otherwise, the trial vector is discarded, as shown in Equation (5).

$$\theta_i^{(t+1)} = \begin{cases} U_i^{(t)}, \text{if } f(U_i^{(t)}) > f(\theta_i^{(t)}) \\ \theta_i^{(t)}, \text{otherwise} \end{cases} \quad (5)$$

where $f(U_i^{(t)})$ and $f(\theta_i^{(t)})$ are the fitness values for $U_i^{(t)}$ and $\theta_i^{(t)}$, respectively.

---

**Algorithm 2**: Differential Evolution

**procedure DE**
    Initialize population as $\Psi_{NP}^{(0)} = \{\theta_1^{(0)}, \theta_2^{(0)}, \dots, \theta_{NP}^{(0)}\}$ with random weights
    $t = 0$
    **while** $f(\theta_{best}) < K$ **and** $t < T_{max}$ **do**
        **for all** $\theta_i \in \Psi_{NP}^{(t)}$ **do**
            $V_i^{(t)} = \theta_{r_1}^{(t)} + F\left(\theta_{r_2}^{(t)} - \theta_{r_3}^{(t)}\right)$
            $U_i^{(t)} = \theta_i^{(t)}$
            **for** $j$ **in** $1: \text{length}\left(U_i^{(t)}\right)$ **do**
                **if** $rand(0,1) \leq CR$ **do**
                      $U_i^{(t)}[j] = V_i^{(t)}[j]$
                **end if**
            **end for**
            **if** $f\left(U_i^{(t)}\right) > f\left(\theta_i^{(t)}\right)$ **do**
                $\theta_i^{(t)} = U_i^{(t)}$
            **end if**
            **if** $f\left(\theta_i^{(t)}\right) > f(p_g)$ **do**
                $p_g = \theta_i^{(t)}$
            **end if**
        **end for**
        $t \mathrel{+}= 1$
    **end while**
**end procedure**

Users can change the operation used to generate the donor vector by setting the scaling factor, $F \in [0,2]$, and selecting from the following operations:

$$\text{DE/rand/1:} \quad V_i^{(t)} = \theta_{r_1}^{(t)} + F\left(\theta_{r_2}^{(t)} - \theta_{r_3}^{(t)}\right) \tag{6}$$

$$\text{DE/rand/2:} \quad V_i^{(t)} = \theta_{r_1}^{(t)} + F\left(\theta_{r_2}^{(t)} - \theta_{r_3}^{(t)}\right) + F\left(\theta_{r_4}^{(t)} - \theta_{r_5}^{(t)}\right) \tag{7}$$

$$\text{DE/best/1:} \quad V_i^{(t)} = p_g + F\left(\theta_{r_1}^{(t)} - \theta_{r_2}^{(t)}\right) \tag{8}$$

$$\text{DE/best/2:} \quad V_i^{(t)} = p_g + F\left(\theta_{r_1}^{(t)} - \theta_{r_2}^{(t)}\right) + F\left(\theta_{r_3}^{(t)} - \theta_{r_4}^{(t)}\right) \tag{9}$$

$$\text{DE/current-to-best:} \quad V_i^{(t)} = \theta_i^{(t)} + F\left(p_g - \theta_{r_1}^{(t)}\right) + F\left(\theta_{r_2}^{(t)} - \theta_{r_3}^{(t)}\right) \tag{10}$$

where $p_g$ represents the neural network with the current best fitness, $r_1$, $r_2$, $r_3$, $r_4$, and $r_5$ are randomly selected distinct integers within the range $[1, NP]$ and are not equal to $i$. The default operation is represented by DE/rand/1.

## 4 Genetic Algorithm

The GA procedure is represented in Algorithm 3. The GA follows closely the standard principles of biological evolution as it cycles through selection, mutation, and recombination. Users can select between tournament selection and fitness proportionate selection. Fitness proportionate selection (or weighted roulette wheel selection) the method depicted in the algorithm below and involves replacing a candidate solution, $\theta_i^{(t)}$, with a probability that is inversely proportional to their fitness. Specifically, $\theta_i^{(t)}$ is replaced with a randomly initialized solution if $rand(0,1) < P\left(\theta_i^{(t)}\right)$ and $P\left(\theta_i^{(t)}\right)$ is calculated as

$$P\left(\theta_i^{(t)}\right) = \frac{f\left(\theta_i^{(t)}\right)}{\sum_{\theta_n \in \Psi_{NP}} f\left(\theta_n^{(t)}\right)}. \tag{11}$$

Tournament selection results in replacing the candidate if its fitness is less than that of a randomly selected individual from the population.

The GA's mutation operation allows for the random alteration of individual in ways that, unlike the recombination mechanisms described for PSO and DE, are not constrained by the current diversity of the population. In this way, the GA is uniquely able to introduce diversity into the population, allowing for individual solutions to emerge that do not resemble any other solution or combination of solutions within the population. Users can select between the mutation operations of random substitution or random interchange. Each element of the target vector is subject to mutation according to the user-provided probability, $p_m \in [0,1]$. In random substitution, mutation involves replacing the element with a random value selected from a uniform distribution defined by the minimum and maximum element values within the current population, $u \sim U(min_{\Psi_{NP}^{(t)}}, max_{\Psi_{NP}^{(t)}})$. In random interchange, the element is switched with a randomly selected element within the solution.

---

**Algorithm 3**: Genetic Algorithm
---
**procedure GA**
    Initialize population as $\Psi_{NP}^{(0)} = \{\theta_1^{(0)}, \theta_2^{(0)}, \dots, \theta_{NP}^{(0)}\}$ with random weights
    $t = 0$
    **while** $f(\theta_{best}) < K$ **and** $t < T_{max}$ **do**
        **for all** $\theta_i \in \Psi_{NP}^{(t)}$ **do**
            **if** $rand(0,1) < P\left(\theta_i^{(t)}\right)$ **do**
                replace $\theta_i^{(t)}$
            **end if**
            **for** $j$ **in** $1: \text{length}\left(\theta_i^{(t)}\right)$ **do**
                **if** $rand(0,1) \leq CR$ **do**
                      $\theta_i^{(t)}[j] = \theta_{r_1}^{(t)}[j]$
                **end if**
                **if** $rand(0,1) \leq p_m$ **do**
                      $\theta_i^{(t)}[j] = u$
                **end if**
            **end for**
        **end for**
        $t += 1$
    **end while**
**end procedure**

## 5  Implementation

The NeuroEvo web platform implements the above algorithms in a cloud-based application that leverages GPU parallelization to improve the speed with which the evolutionary operators are executed across the population. EAs are well-suited for parallelization, as crossover, mutation, and, most importantly, the time-consuming step of evaluating candidate fitness can be performed independently across different individuals in the population. Only steps such as the selection operator for GA, require global information in order to determine the fitness of an individual relative to all others in the current population.

The application is implemented in Python, using PyTorch, and runs on an AWS EC2 p3.8xlarge instance with 4 Tesla V100 GPUs with 16GB of VRAM each. Operators are implemented in a modified version of the Pipe and Filter architecture, as described in [11] and depicted in Figure 1. For the PSO algorithm, the parallelized steps include evaluation of particle fitness, updating particle velocity and position, and, if specified, updating the nonlinear decreasing inertial weight based on the particle's fitness. For DE and GA, the parallelized steps include crossover, mutation, and candidate fitness evaluation and replacement. Upon completion of these steps, the results are aggregated by each algorithm and the stopping criteria are assessed.

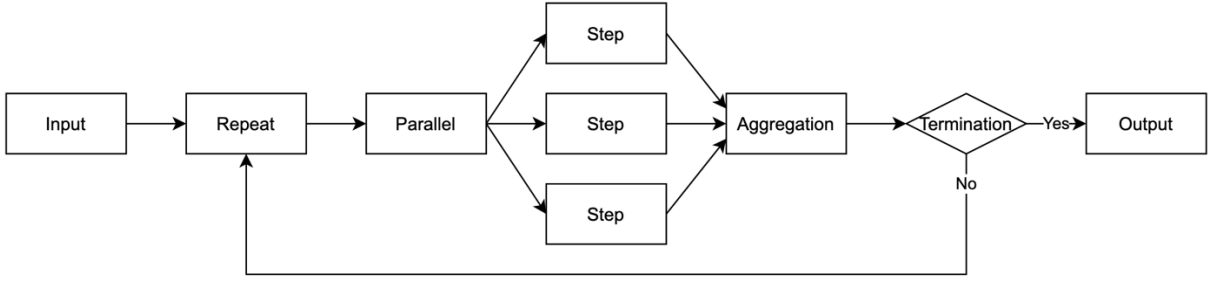

Figure 1: Pipe and Filter architecture

## 6 Neural Network Design and Training

Users can define the classification problem and provide the data used to train the MLP. In addition, users can design the structure of the MLP by selecting the activation function, the number of hidden layers, and the number nodes for each hidden layer, as depicted in Figure 2. The connection between nodes in each layer is defined by weights, $W$, where $W_{j,i}^{(l)}$ represents the weight connecting the output of node $i$ in layer $l$ to node $j$ in layer $l+1$ (starting with $l = 0$ for the layer comprising the input values for each feature). The output of node $i$ in layer $l$ is represented by $a_i^{(l)}$. The output of node $j$ in the first hidden layer ($l = 1$) is calculated as

$$a_i^{(1)} = g\left(\sum_{i=1}^{D} W_{j,i}^{(0)} X_i + W_{1,D+1}^{(0)}\right) \tag{12}$$

where $g(\cdot)$ represents the activation function, $X_i$ represents the input value for feature $i$, $D$ represents the number of nodes in layer $l-1$ (i.e., the number of features), and $W_{1,D+1}^{(0)}$ represents the bias. The output of node $j$ in the hidden layers following the first (i.e., $l > 1$) is then calculated as

$$a_i^{(l)} = g\left(\sum_{i=1}^{D} W_{j,i}^{(l-1)} a_i^{(l-1)} + W_{1,D+1}^{(l-1)}\right). \tag{13}$$

For the activation function, users can select from the following options

$$\text{tanh:} \quad g(x) = \frac{e^x - e^{-x}}{e^x + e^{-x}} \tag{14}$$

$$\text{logistic:} \quad g(x) = \frac{1}{1 + e^x} \tag{15}$$

$$\text{linear:} \quad g(x) = x. \tag{16}$$

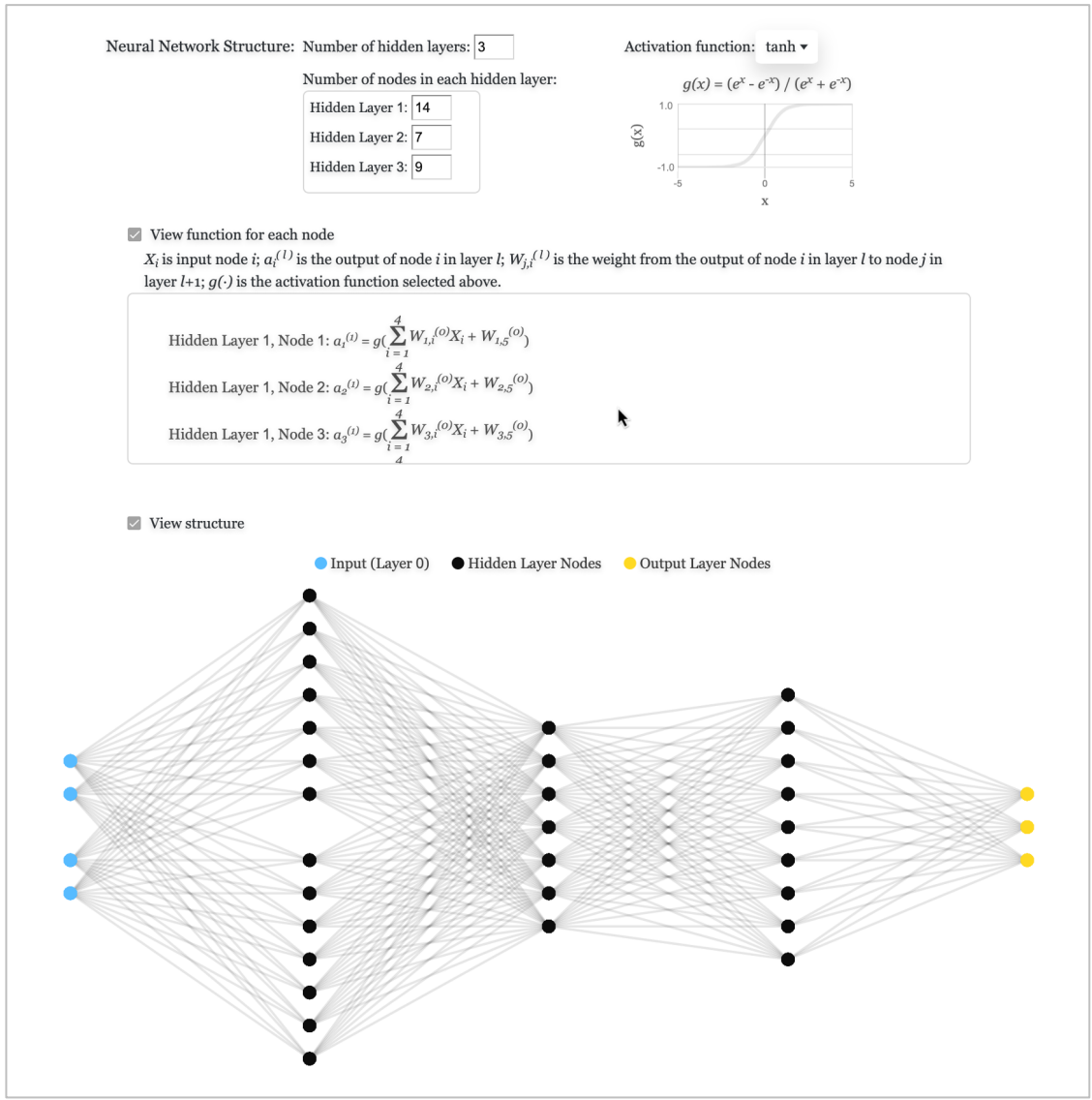

Figure 2: Interface for specifying neural network design

The fitness of the MLP is defined by its classification accuracy. Users can establish stopping criteria for the training process by setting a minimum satisfactory fitness, $K$, for the most-fit MLP in the population, the least-fit MLP in the population, or the average fitness of all MLPs in the population. If this minimum fitness is achieved, the training will be terminated. Otherwise, the training will continue until the maximum number of training iterations, $T_{max}$, is reached.

Users can either upload data to train the MLP or select from example data. Users can also select one of several options for handling missing values and for transforming the raw data. Once the MLP design specifications, training data, and hyperparameters are submitted, the process of input validation will begin. If this concludes without error, the population of candidate MLPs will be initialized and proceed through the chosen training process. Users can view the progression of the population throughout each training iteration. Figure 3 shows an example visualization for PSO (A) and DE (B) of the relative fitness of each candidate, with their position projected according to the first two principal components (PCs), calculated based on the weights of the first generation of candidate solutions.

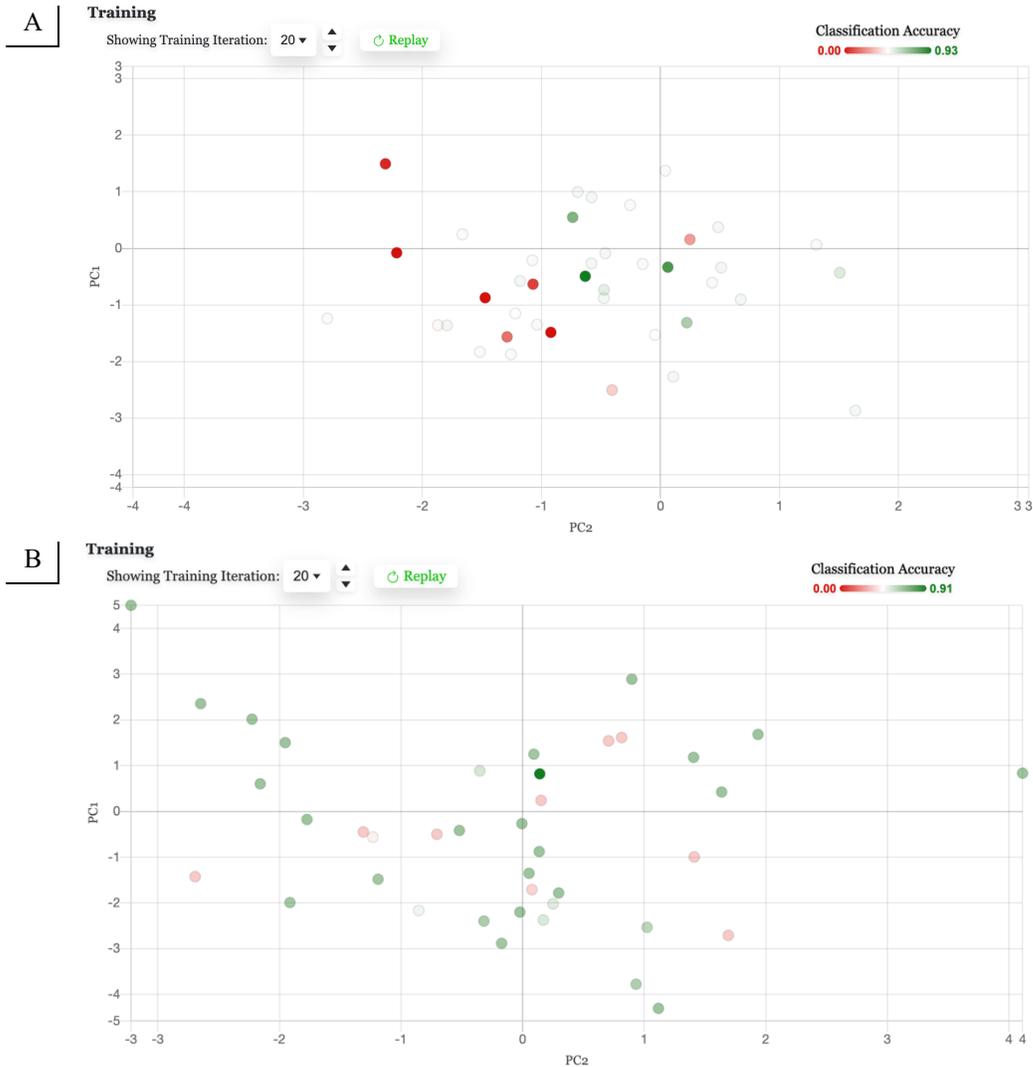

Figure 3: Example view of population training for PSO (A) and DE (B)

Users can also click on any individual candidate to view the trajectory of that candidate as it navigates the search space throughout the training process, as shown in Figure 4 for PSO (A) and DE (B). Figure 4 shows an example of a candidate's trajectory for PSO and DE for 20 training iterations. As described in Algorithm 2, DE only replaces a candidate if it is less fit than the offspring (trial vector). As a result, the candidate is not replaced upon every iteration and there is a monotonic increase in fitness upon every replacement. On the other hand, PSO involves updating the particle's velocity and position upon every iteration, regardless of whether this leads to a decrease or increase in the particle's fitness.

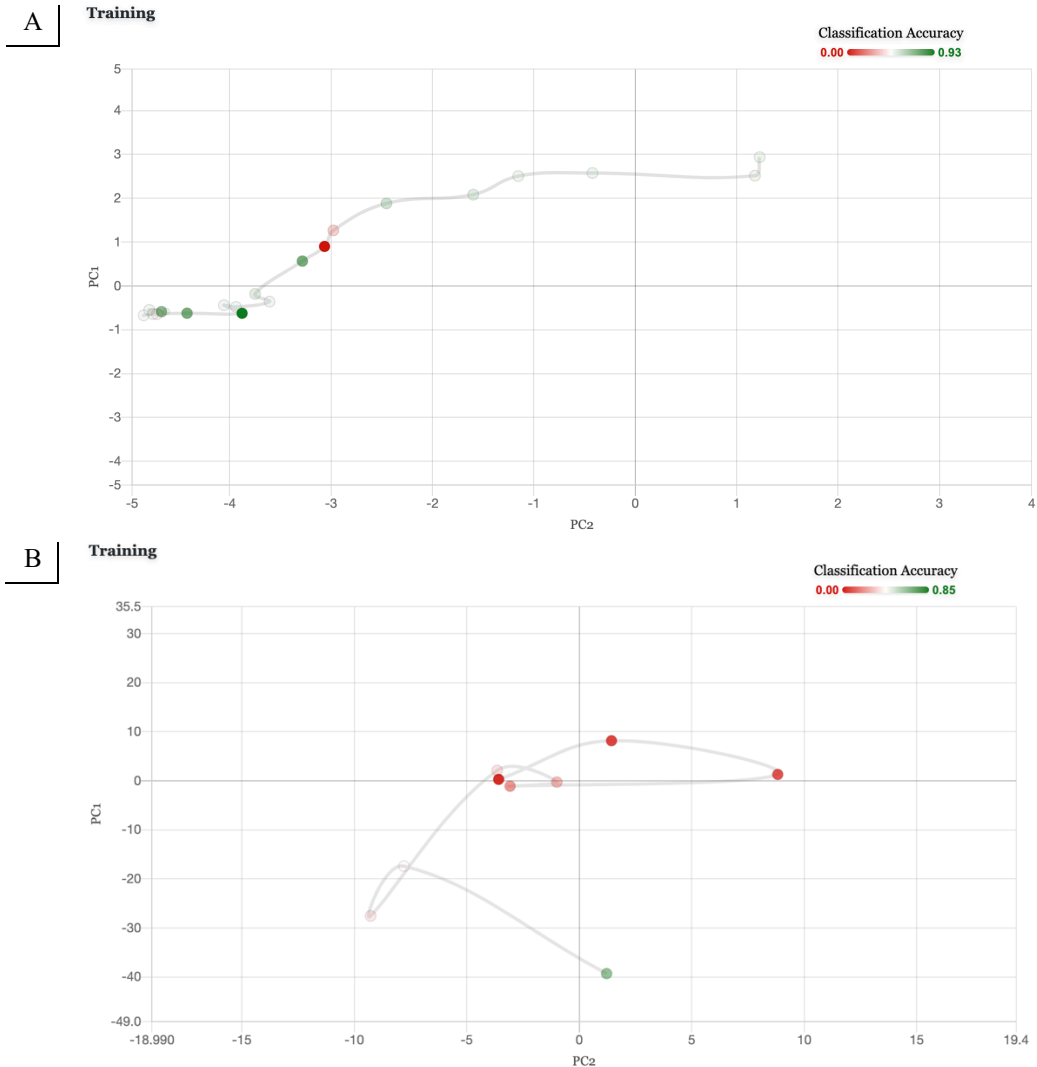

Figure 4: Example view of individual candidate training trajectory for PSO (A) and DE (B)

## 7 Conclusion

NeuroEvo is a computationally efficient cloud-based application that provides an interactive way for users to design and train an MLP classifier using evolutionary and particle swarm algorithms, including PSO, DE, and GA. Users define the classification problem, provide the training data, design the MLP structure, and specify the hyperparameter settings. Upon completion, users can download the trained classifier, provided in Python, Java, and JavaScript, and directly implement it within their environment. The current version of NeuroEvo is to be expanded upon to allow for increased variety in neural network types and increased flexibility in design and training.